\documentclass[final]{cvpr}

\usepackage{times}
\usepackage{epsfig}
\usepackage{graphicx}
\usepackage{amsmath}
\usepackage{amssymb}

\usepackage{booktabs}
\usepackage{algorithm} 
\usepackage{algpseudocode}
\usepackage{mathrsfs}
\usepackage{multirow}
\usepackage{color}
\usepackage{colortbl}
\usepackage{longtable}
\usepackage{colortbl}
\usepackage{amsmath}
\usepackage{amsthm}
\newtheorem{theorem}{Theorem}
\newcommand{\fl}[1]{\lfloor {#1} \rfloor}

\usepackage[pagebackref=true,breaklinks=true,colorlinks,bookmarks=false]{hyperref}

\def\cvprPaperID{7} 
\def\confYear{CVPR 2021}

\begin{document}

\title{AlterSGD: Finding Flat Minima for Continual Learning by Alternative Training$^\clubsuit$}

\author{Zhongzhan Huang\thanks{Equal contribution. \quad $^\clubsuit$ Technical report.}\\
Tsinghua University\\
{\tt\small  zhongzhanhuang@foxmail.com}
\and
Mingfu Liang$^*$\\
Northwestern University\\
{\tt\small  mingfuliang2020@u.northwestern.edu}

\and
Senwei Liang\\
Purdue University\\
{\tt\small liang339@purdue.edu}

\and
Wei He\\
Nanyang Technological University\\
{\tt\small wei005@e.ntu.edu.sg}
}

\maketitle

\begin{abstract}

Deep neural networks suffer from catastrophic forgetting when learning multiple knowledge sequentially, and a growing number of approaches have been proposed to mitigate this problem. Some of these methods achieved considerable performance by associating the flat local minima with forgetting mitigation in continual learning. However, they inevitably need (1)~tedious hyperparameters tuning, and (2)~additional computational cost. To alleviate these problems, in this paper, we propose a simple yet effective optimization method, called AlterSGD, to search for a flat minima in the loss landscape. In AlterSGD, we conduct gradient descent and ascent alternatively when the network tends to converge at each session of learning new knowledge. Moreover, we theoretically prove that such a strategy can encourage the optimization to converge to a flat minima. We verify AlterSGD on continual learning benchmark for semantic segmentation and the empirical results show that we can significantly mitigate the forgetting and outperform the state-of-the-art methods with a large margin under challenging continual learning protocols.  


\end{abstract}

\section{Introduction}
\label{intro}

Recently, deep neural networks~(DNNs) became the dominant paradigm in various real-world applications, for example, image classification~\cite{He_2016_CVPR,huang2020dianet,huang2020efficient,hu2018squeeze,huang2020convolution}, semantic segmentation~\cite{zhang2018context,noh2015learning,long2015fully}, object detection~\cite{viola2001robust,hu2018relation,szegedy2013deep,zhao2019object,redmon2016you}. Typically, the DNN is trained offline by a huge data set sampled independently from the real-world, and it is commonly not feasible when we counter new data which is out-of-distribution with respect to the old data. Therefore, many areas in the Artificial Intelligence seek for the solution by learning new knowledge incrementally without forgetting the learned ones, also known as the continual learning~(CL). 

In this paper, we study the class continual learning~(CCL), where we will incrementally update the DNN to learn a new set of classes disjointed with the learned classes. To reflect the sequential manner of the CCL, we use the term \textit{session} to represent the learning step of each set of classes. For a new session, the corresponding training data is disjoint with all the previous one, which means the new dataset may not have any trained image from previous sessions. However, this is challenging for the DNN since it has intrinsic catastrophic forgetting problem when the DNN is updated incrementally, as pointed in~\cite{kirkpatrick2017overcoming}. In the CCL, the model is trained to learn a new session sequentially with new data which is not seen in previous sessions. At each session, the model needs to adapt to the changes from the new data's distribution, which may potentially lead to the forgetting of the seen class. 

\begin{figure}[t]
    \centering
    \includegraphics[width=1\linewidth]{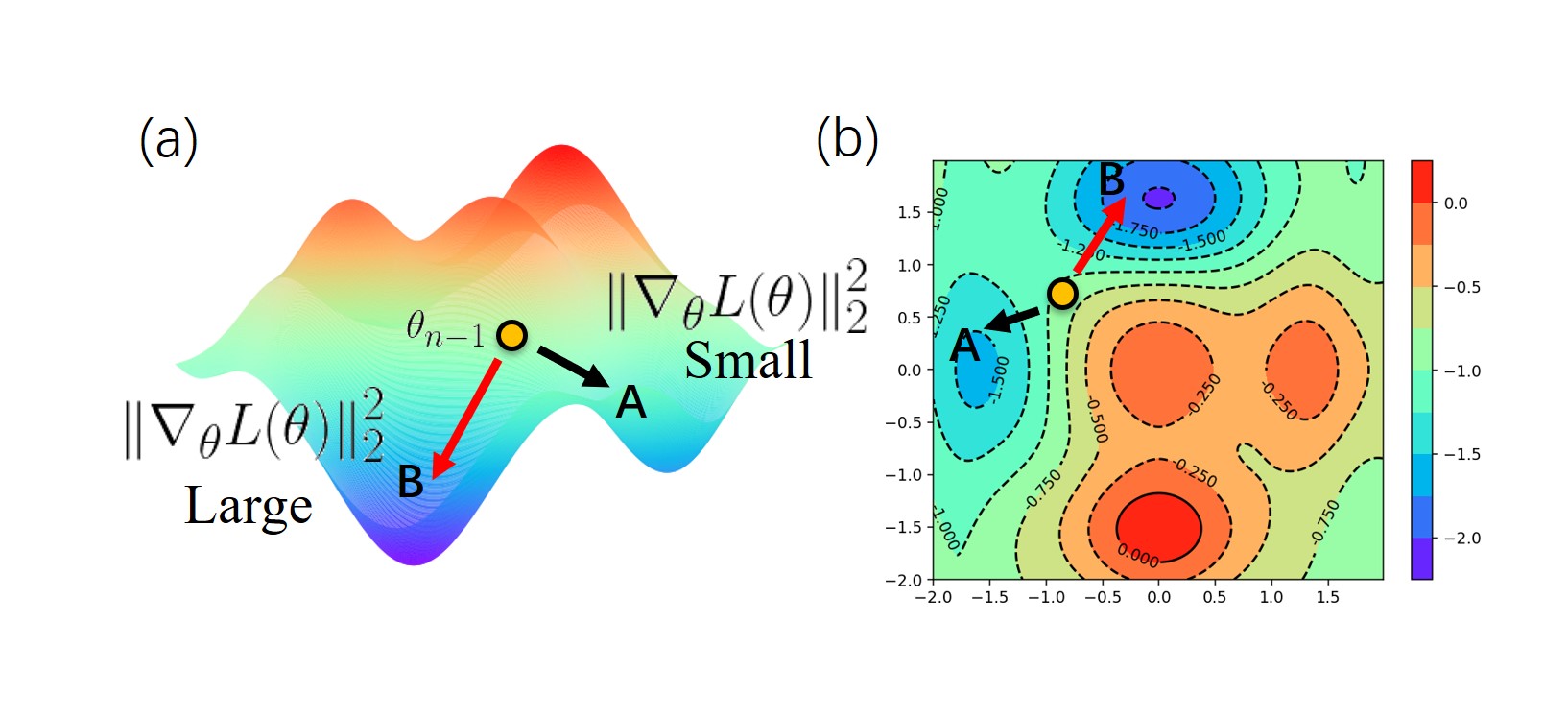}
    \caption{Illustration of AlterSGD. After the normal training of current session tends to converge, AlterSGD conducts gradient descent and ascent alternatively. We show in Theorem~\ref{theo:similarity-layer} that AlterSGD is approximately equal to minimizing $\left\|\nabla_{\theta} L(\theta)\right\|_{2}^{2}$, where $L$ represents the loss function. The smaller gradient norm $\left\|\nabla_{\theta} L(\theta)\right\|_{2}^{2}$ represents flatter the region, e.g., region A has smaller gradient norm than region B. Therefore, AlterSGD tends to converge to a region with a flatter minima, e.g., region A.}
    \label{fig:minimal}
\end{figure}

\begin{figure}[t]
    \centering
    \includegraphics[width=1\linewidth]{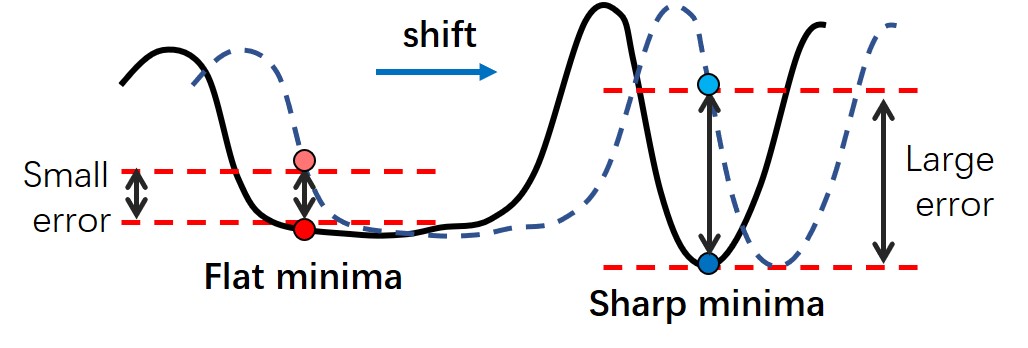}
    \caption{Illustration of loss function shift. The solid curve and the dotted curve represent the loss function over training data of current session and next session respectively. Due to the shift of training data distribution of different sessions, the loss function will shift accordingly. If the training of current session achieves a flat local minima, the training of next session may start from a small error.}
    \label{fig:flataharp}
\end{figure}

On the other hand, in the non-continual learning scenario, \textit{i.e}., learning all the classes at once, we know that when the batch size is too small, the distribution of each batch may extremely vary and cause the shifting of the loss landscape, leading to lousy generalization after training~\cite{keskar2016large,liang2020instance}. One of the solutions for mitigating the distribution variation problem is to seek the flat-minimal region in the loss landscape~\cite{keskar2016large,he2019asymmetric,kleinberg2018alternative,jia2020information}, as illustrated in Figure~\ref{fig:flataharp}. Similarly, in the context of continual learning, such distribution variation will also happen because the training set of current session is disjoint with all the previous data. Hence, we attempt to search for the flat minima in order to potentially benefit the CL. Actually, there are some concurrent works also associating the flat local minimal with forgetting mitigation in CL from different viewpoints (e.g., in CPR~\cite{cha2021cpr} and stable SGD~\cite{NEURIPS2020_518a38cc}). They both verify that maintaining flat local minima can further mitigate the forgetting issue in CL. 
However, these methods have the disadvantages as follows: (1) \textbf{Tedious hyperparameters tuning}. As they may include different regularization methods, it is unavoidable to finely tune the hyperparameter for different vision tasks (e.g. image classification, semantic segmentation) and different continual learning protocols. As pointed out in CPR, they need totally different hyperparameter even for different protocols in the same benchmark dataset (e,g, CIFAR100), which will be laborious and lack of principle to readily make the method work. (2) \textbf{Additional computational cost}. Since the additional regularization methods are existing throughout the new class training, it may inevitably introduce much more computation during training.

Therefore, in this paper, we propose method called AlterSGD: When the network tends to converge at each session, we alternate the gradient direction between descent and ascent such that we can search for the flat local minimal region before training on the next new session. Compared to the existing method, our method is efficient since the alternating of the gradient may not include additional computation and time. Moreover, we theoretically prove that such a strategy can encourage the optimization to converge to a flat minima.

Finally, we verify our method on class continual learning~(CCL) for semantic segmentation and the empirical results show that we can significantly mitigate the forgetting of the classes learned in the initial session, and outperform the state-of-the-art~(SOTA) for a large margin under challenging protocols. To summarize, the contributions of this paper include:
\begin{itemize}
\item We propose a simple and effective method to search for a flat local minimal region such that the forgetting of the learned class can be further mitigated on CCL.
\item We provide the theoretical insight about why the AlterSGD may encourage the optimization to converge to the flat minima. We also conduct extensive experiments to verify the effectiveness of our method on CCL for semantic segmentation.

\end{itemize}

\section{Related Works}

\subsection{Continual Learning}
Aiming to mitigate the \textit{Catastrophic Forgetting} problem, continual learning has been widely discussed on image classification~\cite{kirkpatrick2017overcoming,de2019continual}, which can be divided into replay methods, parameter isolation methods and regularization-based methods~\cite{delange2021continual}. Replay methods~\cite{rebuffi2017icarl,isele2018selective,lopez2017gradient} utilize data samples in the previous session as either the network inputs or the constraints for optimization in the new session. In parameter isolation methods~\cite{xu2018reinforced,mallya2018piggyback}, model parameters are independent for each session, and one can freeze parameters for the previous session to avoid forgetting. Regularization-based methods~\cite{chaudhry2018riemannian,li2017learning,zenke2017continual,kirkpatrick2017overcoming} mostly introduce an extra regularization term in the objective function, so as to strengthen the previously-learned knowledge when training on new session.

Another type of method is promoting wide local minima in continual learning. CPR~\cite{cha2021cpr} proves that maximizing the entropy of the output probability can also improve the accuracy for continual learning. In addition, motivated by the importance of learning stability, stable SGD~\cite{NEURIPS2020_518a38cc} forms the suitable training regimes to widen the local minima in each learning session.

Despite the progress on image classification, recent works further investigate continual learning on object detection~\cite{shmelkov2017incremental} and semantic segmentation~\cite{cermelli2020modeling} with knowledge distillation. In particular, the latter one introduces two novel loss terms to mitigate semantic distribution shift within the background. 

\subsection{Semantic Segmentation}
With the advancement of Fully Convolution Network (FCN) \cite{long2015fully}, recent deep-learning-based approaches on semantic segmentation have achieved impressive progress on several popular benchmarks~\cite{Everingham15,zhou2017scene}. For example, encoder-decoder architectures~\cite{ronneberger2015u, badrinarayanan2017segnet} are the commonly-used and efficient design for the segmentation networks, which share the internal features or information between layers to improve network representation ability. The atrous-convolution-based methods~\cite{zhao2017pyramid, chen2017deeplab,chen2017rethinking} leverage the dilated filters to enlarge the receptive field of the network and utilize different pooling mechanisms~(e.g., atrous spatial pyramid pooling) to acquire richer contextual information. In order to better aggregate semantic context or feature dependencies, other recent works~\cite{YuanW18,fu2020scene} employ self-attention modules on semantic segmentation networks. In addition, to improve the segmentation network efficiency for applications, light-weight design~\cite{paszke2016enet,zhao2018icnet} and network compression~\cite{He_2021_WACV} can also incorporate and boost the advances for this task.

\section{Proposed Method}
In this section, we systematically introduce our proposed AlterSGD for continual learning as outlined in Section~\ref{intro}. The workflow of AlterSGD consists of two phases, the normal training and the alternative training. In the first phase, the model parameters are optimized by normal gradient descent for learning new classes while the second phase is to alternatively conduct gradient descent and ascent such that the optimizer searches for a flat local minima. We summarize our method in Algorithm~\ref{alg:alg}. 

We consider a continual learning problem with $S$ sessions. The empirical loss on the training set at the $s^{\text{th}}$ session is denoted by $L$. We denote $\theta_0$ as the model parameters from the previous session, \textit{i.e.}, $(s-1)^{\text{th}}$ session. Our objective is to learn new classes effectively and simultaneously mitigate forgetting the old classes. Let $T$ be the number of total iterations for parameter update and $p$ be the ratio for normal training. Now we introduce the training details for each phase as follows:

\paragraph{Normal training phase.} In the first phase, the model is trained to learn the new classes at the current session. Therefore, the loss function $L$ is optimized by the normal gradient descent for the first $\fl{pT}$ iterations, \textit{i.e.},
\begin{equation}
    \theta_{t}=\theta_{t-1} - \eta \cdot \nabla_{\theta} L\left(\theta_{t-1}\right),
    \label{eq:AlterSGD}
\end{equation}
where $t=1,2,\cdots, \lfloor pT\rfloor$, $\fl{\cdot}$ denotes the floor function which outputs the greatest integer less than or equal to the input, and $\eta$ is learning rate.

\paragraph{Alternative training phase.} Though gradient descent converges to a local minima for learning a new session after the first phase, it may fall into the sharp minima which possibly aggravate catastrophic forgetting of previous sessions. In the second phase, the proposed AlterSGD utilizes the alternative gradient descent and ascent update rule for $T-\fl{pT}$ iterations, \textit{i.e.},
\begin{equation}
\left\{\begin{array}{l}
\theta_{t}=\theta_{t-1}-\eta \cdot \nabla_{\theta} L\left(\theta_{t-1}\right),\\
\theta_{t+1}=\theta_{t} + \eta \cdot \nabla_{\theta} L\left(\theta_{t}\right),
\end{array}\right.
    \label{eq:AlterSGD}
\end{equation}
where $t=\fl{pT}+1,\fl{pT}+3,\cdots,T$.  

Intuitively, the repeat oscillation by conducting gradient descent and ascent in the parameter space has potential to enable AlterSGD to escape the sharp local minima and approach an equilibrium state where the landscape of the loss function $L$ is relatively flat.
In fact, we show in Theorem~\ref{theo:similarity-layer} that the update rule in Eq.~(\ref{eq:AlterSGD}) is approximately equal to minimizing  $\left\|\nabla_{\theta} L\left(\theta\right)\right\|_2^2$, which means AlterSGD prefers a small norm of $\nabla_{\theta} L\left(\theta\right)$ during alternative training.
\begin{algorithm}[t]  
    \caption{AlterSGD training at the $s^{\text{th}}$ session }\label{alg:alg}   
    \textbf{Input:} The network parameters from the $(s-1)^{\text{th}}$ session, $\theta_0$; Loss function $L(\cdot)$; Learning rate $\eta$; The number of iterations $T$; The alternative ratio $p$.
    
    \textbf{Output:} The network parameters at the $s^{\text{th}}$ session.
        
    \begin{algorithmic}[1]

    \State\algorithmiccomment{Normal Training}

    \For{$t$ from 1 to $\lfloor pT \rfloor$} 
        \State $\theta_{t}=\theta_{t-1}-\eta \cdot \nabla_{\theta} L\left(\theta_{t-1}\right)$
    \EndFor  
    
    \State\algorithmiccomment{Alternative Training}       
        \For{$t=\lfloor pT \rfloor$+1, $\lfloor pT \rfloor+3$, $\cdots, T$} 
        \State $\theta_{t}=\theta_{t-1}-\eta \cdot \nabla_{\theta} L\left(\theta_{t-1}\right)$
        \State $\theta_{t+1}=\theta_{t} {\color{red}{+}} \eta \cdot \nabla_{\theta} L\left(\theta_{t}\right)$
    
    \EndFor
    \State\Return $\theta_T$
    \end{algorithmic} 
\end{algorithm}  

\begin{theorem}\label{theo:similarity-layer}
The update rule of the alternative training in Eq.~(\ref{eq:AlterSGD}) is approximately equal to minimizing the square of $\nabla_{\theta} L\left(\theta\right)$, i.e., $\left\|\nabla_{\theta} L\left(\theta\right)\right\|_2^2$, by gradient descent with learning rate $\frac{\eta^2}{2}$.
\end{theorem}
\begin{proof}
	From the update rule~(\ref{eq:AlterSGD}), we have
	\begin{equation}
	    \begin{aligned}
\theta_{t+1} &=\theta_{t}+\eta \cdot \nabla_{\theta} L\left(\theta_{t}\right) \\
&=\theta_{t-1}-\eta \nabla_{\theta} L\left(\theta_{t-1}\right)+\eta\nabla_{\theta} L\left(\theta_{t-1}-\eta \nabla_{\theta} L\left(\theta_{t-1}\right)\right).
\end{aligned}
	\end{equation}
	Since $\eta$ is small, $\nabla_{\theta} L\left(\theta_{t-1}-\eta \nabla_{\theta} L\left(\theta_{t-1}\right)\right)$ can be approximated by its Taylor expansion. Then we have 
\begin{equation}
\begin{aligned}
\theta_{t+1} \approx \theta_{t-1} &-\eta \nabla_{\theta} L\left(\theta_{t-1}\right)+\eta\nabla_{\theta} L\left(\theta_{t-1}\right) \\
&-\eta \cdot \eta \cdot \nabla_{\theta}^{2} L\left(\theta_{t-1}\right)^{\top} \nabla_{\theta} L\left(\theta_{t-1}\right).\\
\end{aligned}
\label{eqn:taylor}
    \end{equation}
By~\eqref{eqn:taylor}, 
$\theta_{t+1} \approx \theta_{t-1}-\frac{\eta^{2}}{2} \nabla_{\theta}\left\|\nabla_{\theta} L\left(\theta_{t-1}\right)\right\|^{2}$, which indicates the update rule~(\ref{eq:AlterSGD}) is approximately equal to minimizing  $\left\|\nabla_{\theta} L\left(\theta\right)\right\|_2^2$.

	\qedhere
\end{proof}

Now we illustrate the behavior of AlterSGD using the conclusion of Theorem~\ref{theo:similarity-layer} and how it is related to our goal: As shown in Figure~\ref{fig:minimal}, both region A and B have a local minima but the local minima in region A is flatter than that in region B. Also, since the slope of region B is steeper than region A, the norm of gradient in region A is smaller than that in region B. Finally, AlterSGD tends to converge to region A, a flatter minima, where the norm of gradient is smaller.

\begin{table*}[htbp]
  \centering
  \small
  \caption{Per class mean IoU on 15-1 setting of PASCAL VOC for the first 15 classes for different continual learning methods. }
    \begin{tabular}{c|ccccccccccccccc}
    \toprule
    \textbf{Method} & aero  & bike  & bird  & boat  & bottle & bus   & car   & cat   & chair & cow   & table & dog   & horse & mbike & person \\
    \midrule
    FT    & 0.3   & 0.0   & 0.0   & 2.5   & 0.0   & 0.0   & 0.0   & 0.0   & 0.0   & 0.0   & 0.0   & 0.0   & 0.0   & 0.0   & 0.0 \\
    PI    & 0.0   & 0.0   & 0.0   & 0.0   & 0.0   & 0.0   & 0.0   & 0.0   & 0.0   & 0.0   & 0.0   & 0.0   & 0.0   & 0.0   & 0.0 \\
    EWC   & 0.0   & 0.0   & 0.0   & 1.0   & 0.0   & 0.0   & 0.0   & 0.0   & 0.0   & 0.0   & 3.6   & 0.0   & 0.0   & 0.0   & 0.0 \\
    RW    & 0.0   & 0.0   & 0.0   & 0.2   & 0.0   & 0.0   & 0.0   & 0.0   & 0.0   & 0.0   & 2.2   & 0.0   & 0.0   & 0.0   & 0.0 \\
    LwF   & 0.0   & 0.0   & 0.0   & 0.0   & 0.6   & 0.0   & 0.0   & 0.0   & 0.0   & 0.0   & 0.0   & 0.0   & 0.0   & 0.0   & 0.0 \\
    LwF-MC & 0.0   & 6.3   & 0.8   & 0.0   & 1.1   & 0.0   & 0.0   & 0.0   & 0.0   & 0.0   & 0.0   & 0.0   & 0.0   & 0.0   & 0.0 \\
    ILT   & 3.7   & 0.0   & 2.9   & 0.0   & 12.8  & 0.0   & 0.0   & 0.1   & 0.0   & 0.0   & \textbf{21.2} & 0.1   & 0.4   & 0.6   & 13.6 \\
    \midrule
    MiB   & 53.8  & 38.7  & 35.5  & 25.9  & 45.3  & 14.2  & 71.5  & 71.5  & 0.3   & 22.6  & 12.6  & 62.6  & 52.3  & 48.8  & \textbf{79.8} \\
    Ours  & \textbf{64.6} & \textbf{39.4} & \textbf{44.5} & \textbf{39.1} & \textbf{56.4} & \textbf{57.4} & \textbf{78.5} & \textbf{74.5} & \textbf{2.4} & \textbf{51.7} & 14.2  & \textbf{66.2} & \textbf{60.2} & \textbf{62.5} & 78.6 \\
    \midrule
    Joint & 90.2  & 42.2  & 89.5  & 69.1  & 82.3  & 92.5  & 90.0  & 94.2  & 39.2  & 87.6  & 56.4  & 91.2  & 86.8  & 88.0  & 86.8 \\
    \bottomrule
    \end{tabular}%
  \label{tab:1-15}%
\end{table*}%
\begin{table}[htbp]
  \centering
  \small
  \caption{Per class mean IoU for the last 5 classes.}
  \resizebox{\columnwidth}{!}{%
    \begin{tabular}{c|c|c|c|c|c|c|c|c}
    \toprule
    \textbf{Method} & plant & sheep & sofa  & train & tv    & \textbf{1-15} & \textbf{16-20} & \textbf{all} \\
    \midrule
    FT    & 0.0   & 0.0   & 0.0   & 0.0   & 8.8   & 0.2   & 1.8   & 0.6 \\
    PI    & 0.0   & 0.0   & 0.0   & 0.3   & 8.6   & 0.0   & 1.8   & 0.4 \\
    EWC   & 0.0   & 0.0   & 7.3   & 7.0   & 7.4   & 0.3   & 4.3   & 1.3 \\
    RW    & 0.0   & 0.0   & 8.1   & 10.5  & 8.2   & 0.2   & 5.4   & 1.5 \\
    LwF   & 0.0   & 0.0   & 1.9   & 8.2   & 7.9   & 0.8   & 3.6   & 1.5 \\
    LwF-MC & 0.0   & 0.0   & 2.9   & 11.9  & 11.0  & 4.5   & 7.0   & 5.2 \\
    ILT   & 0.0   & 0.0   & 11.6  & 8.3   & 8.5   & 3.7   & 5.7   & 4.2 \\
    \midrule
    MiB   & \textbf{10.9} & 13.3  & \textbf{10.7} & 18.1  & 10.1  & 42.4  & 12.6  & 34.9 \\
    Ours  & 8.4   & \textbf{13.6} & 8.8   & \textbf{22.9} & \textbf{10.7} & \textbf{52.7} & \textbf{12.9} & \textbf{42.7} \\
    \midrule
    Joint & 62.3  & 88.4  & 49.5  & 85.0  & 78.0  & 79.1  & 72.6  & 77.4 \\
    \bottomrule
    \end{tabular}%
    }
  \label{tab:15-20}%
\end{table}%

\section{Experiments}
In this section, we demonstrate the effectiveness of our proposed AlterSGD as presented in Algorithm~\ref{alg:alg}. The performance is evaluated on the continual class learning~(CCL) in semantic segmentation. 
To compare with popular methods on CCL, we use the baselines as suggested in \cite{michieli2019incremental}, including Elastic Weight
Consolidation~(EWC)~\cite{kirkpatrick2017overcoming}, Path Integral~(PI)~\cite{zenke2017continual}, Riemannian Walks~(RW)~\cite{chaudhry2018riemannian}, Learning without Forgetting~(LwF)~\cite{li2017learning}, LwF-MC~\cite{rebuffi2017icarl}, ILT\cite{michieli2019incremental}, and MiB~\cite{cermelli2020modeling}. To the best of our knowledge, MiB is the state-of-the-art~(SOTA) technique on CCL in semantic segmentation, and we show that AlterSGD can even outperform MiB. 

The regularization-based methods~\cite{kirkpatrick2017overcoming,zenke2017continual,chaudhry2018riemannian,michieli2019incremental,cermelli2020modeling} mitigate forgetting by imposing a regularization to restrain the model parameters. At the $s^\text{th}$ session, we denote $D^{s}$ as the training set with $|D^{s}|$ samples and $\theta^s$ as the model parameters. The objective function of regularization-based methods can be formulated as, 
\begin{equation}
    \frac{1}{\left|D^{s}\right|} \sum_{\left(x, y^{g t}\right) \in D^{s}}\left(\ell_{s e g}^{\theta^{s}}\left(x, y^{g t}\right)+\lambda_{reg} R^{\theta^s}(x) \right),
    \label{eq:ss_cl}
\end{equation}
where $\ell_{s e g}^{\theta^{s}}\left(\cdot \right)$ and $R^{\theta^s}(\cdot)$ represent the segmentation loss and
the regularization term respectively, $y^{gt}$ is the ground-truth segmentation map of the input image $x$, and 
$\lambda_{reg}$ is the regularization coefficient. In our experiments, we apply AlterSGD to Eq.~(\ref{eq:ss_cl}) on the supervised loss term $\ell_{s e g}^{\theta^{s}}\left(\cdot \right)$, which utilizes the update rule as follows,
\begin{equation}
    \left\{\begin{array}{l}
\theta_{t}^s=\theta_{t-1}^s-\eta \cdot \nabla_{\theta} \big[\ell_{s e g}^{\theta}\left(\theta_{t-1}^s\right) + \lambda_{reg} R^{\theta}\left(\theta_{t-1}^s\right) \big], \\
\theta_{t+1}^s=\theta_{t}^s-\eta \cdot \nabla_{\theta} [{\color{red}{-}}\ell_{s e g}^{\theta}\left(\theta_{t}^s\right) + \lambda_{reg} R^{\theta}\left(\theta_{t}^s\right) ]
\end{array}\right.
\label{eq:6}
\end{equation}
where $\ell_{s e g}^{\theta}\left(\theta_{t}^s\right) = \frac{1}{\left|D^{s}\right|} \sum_{\left(x, y^{g t}\right) \in D^{s}}\ell_{s e g}^{\theta_{t}^s}\left(x, y^{g t}\right)$, and $R^{\theta}\left(\theta_{t}^s\right) = \frac{1}{\left|D^{s}\right|} \sum_{\left(x, y^{g t}\right) \in D^{s}}R^{\theta_{t}^s}(x)$.

\paragraph{Dataset \& settings.}
We evaluate the performance on the benchmark dataset for semantic segmentation, PASCAL VOC 2012~\cite{Everingham15}. PASCAL VOC contains 20 foreground semantic classes and a background class. 
For each experiment, we report the mean Intersection-over-Union~(mIoU) in percentage on the validation set. The CCL experiments are conducted on 15-1 setting where the model is trained on first 15 classes in the initial session and then trained on another 5 classes sequentially. For each session, we only provide the label of the classes we will learn in the current session and the learned classes from the previous sessions will be masked as background. Following the settings in~\cite{cermelli2020modeling}, in the initial session, the model is trained for 30 epochs with initial learning rate 0.01 to well-train the 15 classes; For the 5 continual sessions, the learning rate is changed to 0.001 and the model is still trained for 30 epochs in each session. We choose $\lambda_{reg}$ to be 100 according to~\cite{cermelli2020modeling} and choose $p=25/30$. As the old data is not allowed to be accessed during new classes training, therefore we do not consider the replay method for comparison as they may store the old data.

\begin{figure*}
    \centering
    \includegraphics[width=0.9\linewidth]{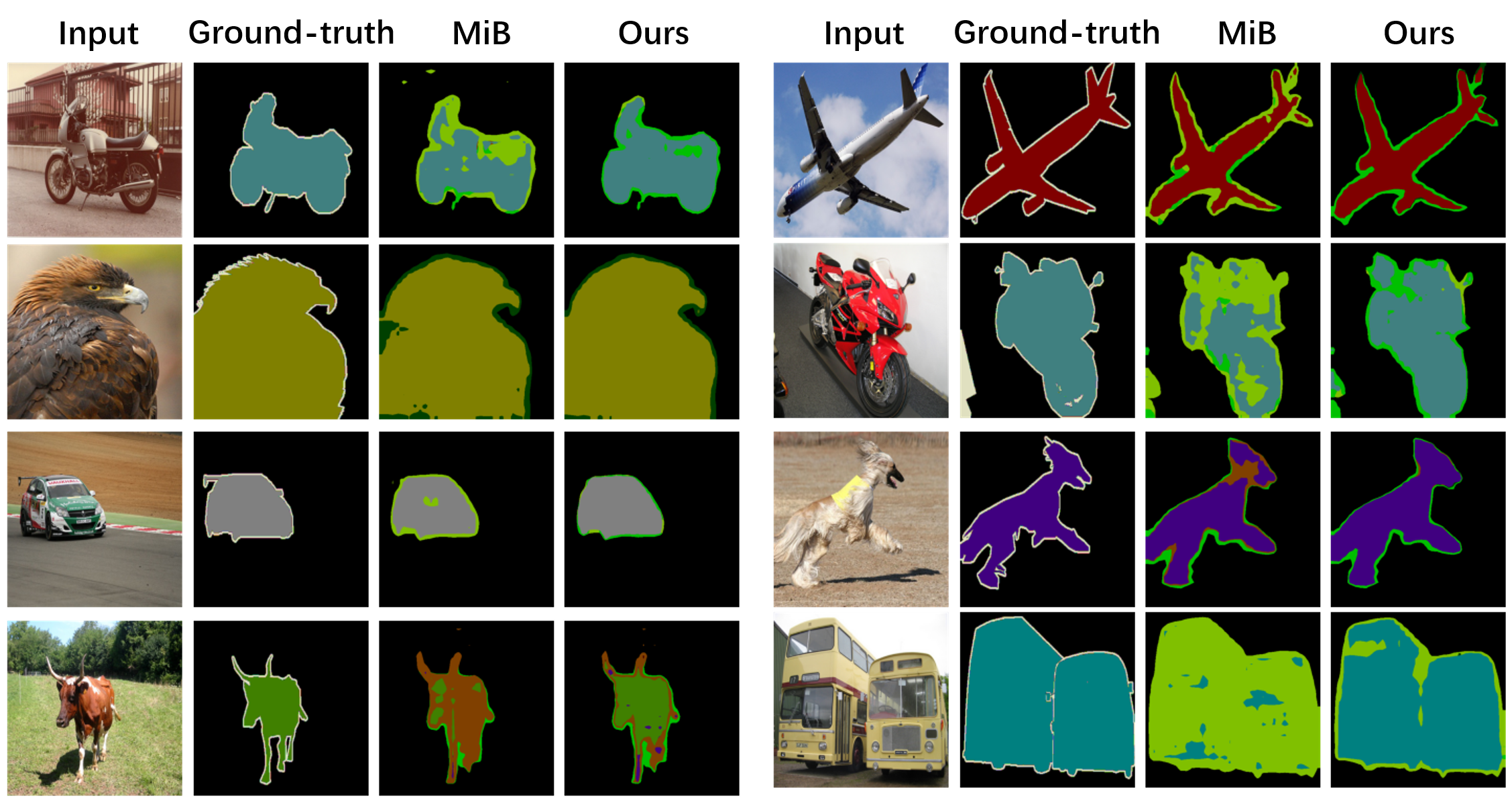}
    \caption{Visualization results on the 15-1 setting of PASCAL VOC test set using MiB and AlterSGD (ours).}
    \label{fig:seg}
\end{figure*}

\paragraph{Experimental results.}
Table~\ref{tab:1-15} displays the result for the first 15 classes using different methods, and Table~\ref{tab:15-20} presents the result of the last 5 classes and the overall results. To fairly compare our method with MiB~\cite{cermelli2020modeling}, we use the same pretrained model obtained from the initial session, and use it to continually learn for five sessions under MiB~\cite{cermelli2020modeling} and our method respectively, reported in Table~\ref{tab:1-15} and~\ref{tab:15-20}.
In Table~\ref{tab:1-15} and Table~\ref{tab:15-20}, the regularization-based methods, like EWC~\cite{kirkpatrick2017overcoming}, PI~\cite{zenke2017continual}, RW~\cite{chaudhry2018riemannian}, LwF~\cite{li2017learning}, LwF-MC~\cite{rebuffi2017icarl}, ILT\cite{michieli2019incremental} obtain the extremely poor results. The mIoU of both the first 15 classes and the last 5 classes are small~(closed to zero in most classes), which means that these methods are not directly effective in the continual learning for semantic segmentation, though they perform  well on image classification. 

Compared with MiB, AlterSGD achieves comparable results on the newly learned classes. Moreover, AlterSGD significantly outperforms MiB on the first 1-15 classes. On average over the first 15 classes, we improve the mIoU of MiB from 42.4 to 52.7, which means AlterSGD is capable of mitigating forgetting effectively. To sum up, the proposed AlterSGD is able to achieve satisfactory results on both newly learned session and mitigating forgetting. 

In Figure~\ref{fig:seg}, we show a qualitative analysis through visualizing the predicted segmentation map of the test images under MiB and our method respectively. Although MiB is the SOTA technique on CCL in semantic segmentation, from the visualization results we can observe that there is still a large segmentation error compared with ground-truth image, while this error can be significantly alleviated by the AlterSGD.

\begin{table}[tbp]
  \centering
  \small
  \caption{Mean IoUs of PASCAL VOC for different regularization coefficient $\lambda_{reg}$. }
    \begin{tabular}{c|cc|cc|cc}
    \toprule
          & \multicolumn{2}{c|}{1-15} & \multicolumn{2}{c|}{16-20} & \multicolumn{2}{c}{all} \\
\cmidrule{2-7}    $\lambda_{reg}$   & MiB   & Ours  & MiB   & Ours  & MiB   & Ours \\
    \midrule
    $10^0$     & 30.9  & \textbf{36.1} & 11.4  & \textbf{11.8}  & 26.1  & \textbf{30.0}~({\color{red}{$\uparrow$ 3.9}}) \\
    $10^1$    & 36.3  & \textbf{48.9}  & \textbf{13.9}  & 13.7  & 30.7  & \textbf{40.1}~({\color{red}{$\uparrow$ 9.4}}) \\
    $10^2$   & 42.4  & \textbf{52.7}  & 12.6  & \textbf{12.9 } & 34.9  & \textbf{42.7}~({\color{red}{$\uparrow$ 7.8}}) \\
    $10^3$  & 16.8  & \textbf{19.5}  & \textbf{5.6}   & \textbf{5.6}   & 13.9  & \textbf{16.0}~({\color{red}{$\uparrow$ 2.1}}) \\
    \bottomrule
    \end{tabular}%
  \label{tab:diffreg}%
\end{table}%

\begin{table*}[t]
  \centering
  \caption{Mean IoUs of PASCAL VOC for different regularization coefficient combination $(\lambda_a, \lambda_b)$.}
    \begin{tabular}{c|cccc|cccc|cccc}
    \toprule
          & \multicolumn{4}{c|}{\textbf{1-15}} & \multicolumn{4}{c|}{\textbf{16-20}} & \multicolumn{4}{c}{\textbf{all}} \\
    \midrule
     $\lambda_b \backslash \lambda_a$  & $10^0$     & $10^1$    & $10^2$   & $10^3$  & $10^0$     & $10^1$    & $10^2$   & $10^3$  & $10^0$     & $10^1$    & $10^2$   & $10^3$ \\
          \midrule
    $10^0$     & 36.1  & 45.2  & \cellcolor[rgb]{ 1,  .114,  .114}\textbf{55.1} & 18.4  & 11.8  & \cellcolor[rgb]{ 1,  .294,  .294}\textbf{14.4} & 9.5   & 3.9   & 30.0  & 37.5  & \cellcolor[rgb]{ 1,  .294,  .294}\textbf{43.7} & 14.8 \\
    $10^1$    & 46.7  & 48.9  & \cellcolor[rgb]{ 1,  .294,  .294}\textbf{55.0} & 17.7  & \cellcolor[rgb]{ 1,  .788,  .788}\textbf{11.9} & \cellcolor[rgb]{ 1,  .506,  .506}\textbf{13.7} & 9.7   & 4.2   & 38.0  & 40.1  & \cellcolor[rgb]{ 1,  .294,  .294}\textbf{43.7} & 14.4 \\
    $10^2$   & 46.2  & \cellcolor[rgb]{ 1,  .506,  .506}\textbf{53.2} & \cellcolor[rgb]{ 1,  .655,  .655}\textbf{52.7} & 17.2  & 9.6   & \cellcolor[rgb]{ 1,  .114,  .114}\textbf{14.8} & \cellcolor[rgb]{ 1,  .655,  .655}\textbf{12.9} & 4.9   & 37.1  & \cellcolor[rgb]{ 1,  .506,  .506}\textbf{43.6} & \cellcolor[rgb]{ 1,  .655,  .655}\textbf{42.7} & 14.1 \\
    $10^3$  & 42.3  & \cellcolor[rgb]{ 1,  .788,  .788}\textbf{51.1} & 49.0  & 19.5  & 2.3   & 8.5   & 13.7  & 5.6   & 32.3  & \cellcolor[rgb]{ 1,  .788,  .788}\textbf{40.5} & 40.2  & 16.0 \\
    \bottomrule
    \end{tabular}%
  \label{tab:reg combi}%
\end{table*}%

\section{Analysis}
\subsection{Varying the alternative ratio $p$}
In Algorithm~\ref{alg:alg}, we use the hyperparamter $p$ to control the ratio of the length of normal training in each session. To study the appropriate $p$ for AlterSGD, we conduct CCL on 15-1 setting of PASCAL VOC for different ratio $p$. The number of epoch is 30 for each session and we present the results of $p=5/30,10/30,\cdots,30/30$ for conducting alternative training phase after the $5, 10, \cdots, 30$ epoch respectively. When $p=30/30$, AlterSGD only contains the normal training phase.

\begin{figure}
    \centering
    \includegraphics[width=1\linewidth]{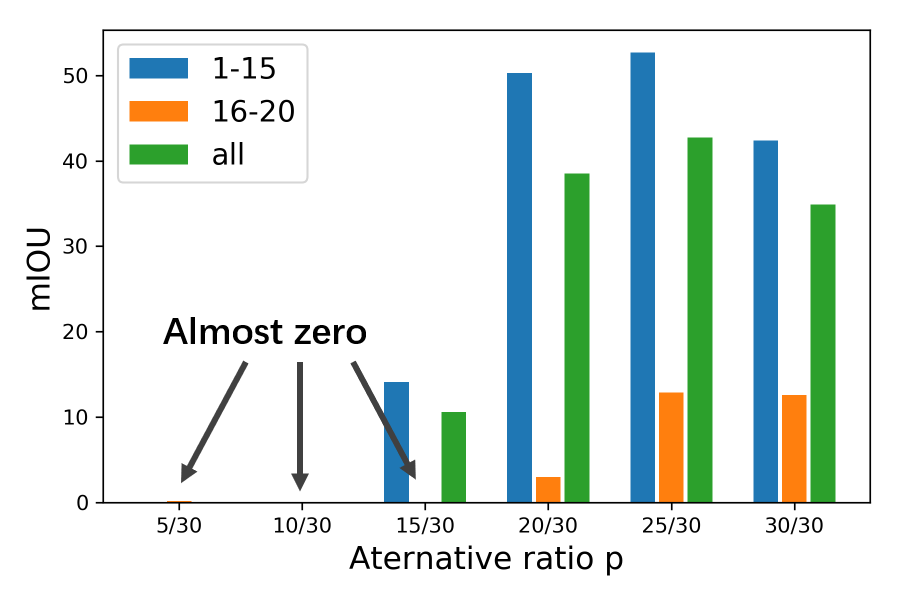}
    \caption{Mean IoU of the first 15 classes (1-15), last 5 classes (16-20), and all 20 classes (all) for different alternative ratio $p$. We use $p$ to denote the ratio of the length of normal training.}
    \label{fig:p}
\end{figure}

When $p$ is small (like $p=5/30,10/30$), the mIoUs of all 20 classes are almost zero. Besides, when $p = 15/30$, though mIoUs of the first 15 classes are 14.1, the mIoUs of the last 5 classes are still almost 0 (see Figure~\ref{fig:p}). 

These phenomena reveal that $p$ should not be too small. This is because when $p$ is small, on one hand, the model does not have enough epochs to train the model in the normal training phase, which is not conducive to the learning of the current class. On the other hand, according to Theorem \ref{theo:similarity-layer}, the optimizer almost always searches for a relatively flat local minima, which affects finding a small enough local minima for the current class. 
Therefore, we suggest to start alternative training phase only after sufficient normal training, \textit{i.e.}, the training of the current class is close to convergence. Generally, $p$ can be selected in $[0.8,0.9]$.

\subsection{Varying the penalty coefficient $\lambda_{reg}$}
\label{sec:reg influence}
 The regularization items $R^{\theta^s}(\cdot)$ in \eqref{eq:6} are usually designed to maintain the performance of the old classes. The penalty coefficient $\lambda_{reg}$ represents the intensity of the penalty, which has a great impact on the performance of regularization methods for continual learning~\cite{hsu2018re}. In this section, we study the influence of $\lambda_{reg}$ in AlterSGD. Table~\ref{tab:diffreg} displays results of $\lambda_{reg}$ of different magnitude.

Since AlterSGD is applied to MiB, the trend of performance change of AlterSGD and MiB presented in Table~\ref{tab:diffreg} is consistent, \textit{i.e.}, the large mIoU of MiB, the larger mIoU of AlterSGD. Further, under different choose of $\lambda_{reg}$, AlterSGD can consistently improve the performance. Specifically, AlterSGD significantly outperforms MiB on the first 15 classes and maintains the performance of last 5 classes.

\subsection{Varying $\lambda_{reg}$ in the training phase}
In our original setting in Eq.~\ref{eq:6}, $\lambda_{reg}$ is the same for gradient descent and ascent. In this section, we explore the impact of using separately different $\lambda_{reg}$ for gradient ascent and gradient descent. Eq.~(\ref{eq:6}) can be rewritten as

\begin{equation}
    \left\{\begin{array}{l}
\theta_{t}^s=\theta_{t-1}^s-\eta \cdot \nabla_{\theta} \big[\ell_{s e g}^{\theta}\left(\theta_{t-1}^s\right) + {\color{red}{\lambda_{a}}} R^{\theta}\left(\theta_{t-1}^s\right) \big], \\
\theta_{t+1}^s=\theta_{t}^s-\eta \cdot \nabla_{\theta} [-\ell_{s e g}^{\theta}\left(\theta_{t}^s\right) + {\color{red}{\lambda_{b}}} R^{\theta}\left(\theta_{t}^s\right) ]
\end{array}\right.
\end{equation}
Table~\ref{tab:reg combi} shows the results of different combination of $\lambda_a$ and $\lambda_b$. 
The different combination of $(\lambda_{a},\lambda_b)$ for gradient ascent and gradient descent can indeed bring some performance improvements. However, compared with Table~\ref{tab:diffreg} where $\lambda_a = \lambda_b$, the improvements are marginal. It suggests that we can keep the same $\lambda_{reg}$ in AlterSGD when using gradient ascent and gradient descent alternatively.

\section{Conclusion}
In this paper, we study the class continual learning on semantic segmentation under the setting that the old data is inaccessible. In particular, we propose a simple and effective method to search for a flat local minimal region such that the forgetting problem of the learned class can be mitigated during the learning of new session. Moreover, we provide the theoretical insight to illustrate that our method can encourage the optimization to converge toward the flat minima. The extensive experiments verify the effectiveness of our method on CCL for semantic segmentation.


\end{document}